\def\year{2021}\relax
\documentclass[letterpaper]{article} % DO NOT CHANGE THIS
\usepackage{aaai21}  % DO NOT CHANGE THIS
\usepackage{times}  % DO NOT CHANGE THIS
\usepackage{helvet} % DO NOT CHANGE THIS
\usepackage{courier}  % DO NOT CHANGE THIS
\usepackage[hyphens]{url}  % DO NOT CHANGE THIS
\usepackage{graphicx} % DO NOT CHANGE THIS
\usepackage{natbib}  % DO NOT CHANGE THIS AND DO NOT ADD ANY OPTIONS TO IT
\usepackage{caption} % DO NOT CHANGE THIS AND DO NOT ADD ANY OPTIONS TO IT
 \usepackage{booktabs}       % professional-quality tables
\usepackage{subcaption}
\usepackage{siunitx}

\usepackage{graphicx}
\usepackage{amsmath}
\usepackage{amssymb}
\usepackage{amsthm}
\usepackage{bbold}
\usepackage{nth}
\usepackage{makecell}

\title{Illuminating Mario Scenes in the Latent Space \\ of a Generative Adversarial Network}
\author {
    Matthew C. Fontaine\textsuperscript{\rm 1},
    Ruilin Liu\textsuperscript{\rm 1},
    Ahmed Khalifa\textsuperscript{\rm 2}, 
    Jignesh Modi\textsuperscript{\rm 1}, \\
    Julian Togelius\textsuperscript{\rm 2},
     Amy K. Hoover\textsuperscript{\rm 3},
     Stefanos Nikolaidis\textsuperscript{\rm 1} \\
}

\affiliations {
    \textsuperscript{\rm 1} University of Southern California \\
    \textsuperscript{\rm 2} New York University \\
    \textsuperscript{\rm 3} New Jersey Institute of Technology \\
    mfontain@usc.edu, ruilinli@usc.edu, aak538@nyu.edu, jigneshm@usc.edu, \\ 
    julian@togelius.com, ahoover@njit.edu, nikolaid@usc.edu
}

\begin{document}

\maketitle

\begin{abstract}

Generative adversarial networks (GANs) are quickly becoming a ubiquitous approach to procedurally generating video game levels. While GAN generated levels are stylistically similar to human-authored examples, human designers often want to explore the generative design space of GANs to extract interesting levels. However, human designers find latent vectors opaque and would rather explore along dimensions the designer specifies, such as number of enemies or obstacles. We propose using state-of-the-art quality diversity algorithms designed to optimize continuous spaces, i.e. MAP-Elites with a directional variation operator and Covariance Matrix Adaptation MAP-Elites, to efficiently explore the latent space of a GAN to extract levels that vary across a set of specified gameplay measures. In the benchmark domain of Super Mario Bros, we demonstrate how designers may specify gameplay measures to our system and extract high-quality (playable) levels with a diverse range of level mechanics, while still maintaining stylistic similarity to human authored examples. An online user study shows how the different mechanics of the automatically generated levels affect subjective ratings of their perceived difficulty and appearance.

% Old
%Recent developments in machine learning techniques, such as generative adversarial networks (GANs), have allowed automatic generation of video game levels that are stylistically similar to human-authored examples. However, a designer may wish to generate not just any level that resembles a training dataset, but a set of high quality, playable levels that exhibit a diverse range of desired characteristics, such as number of enemies. We propose using state-of-the-art quality diversity algorithms designed to optimize in continuous spaces, i.e., MAP-Elites with a directional variation operator and Covariance Matrix Adaptation MAP-Elites, to efficiently search the parameter space of the GAN along a set of multiple level mechanics. In the benchmark domain of Super Mario Bros, we show that this allows us to search the latent variables of a GAN to generate high-quality (playable) levels with a diverse range of prespecified level mechanics, while still maintaining stylistic similarity to human examples. An online user study shows how the different mechanics of the automatically generated levels affect subjective ratings of their perceived difficulty and appearance.

\end{abstract}

%\keywords{Quality diversity, illumination algorithms, generative adversarial network, Mario, MAP-Elites, CMA-ME}  % put your semicolon-separated keywords here!

%\maketitle
\section{Introduction}

Algorithms that procedurally generate content often need to adhere to a desired style or aesthetics. For example, generative adversarial networks (GANs)~\citep{goodfellow:nips14, karras:iclr2018} generate realistic looking images after training on a large dataset of human specified examples. At the same time, for these algorithms to be useful in practice, they need to enable generation of a \textit{diverse} range of content, across a range of attributes specified by a human designer. For a GAN, this requires either sifting through thousands of randomly generated examples, which is cost-prohibitive, or controlling the GAN output by ``steering'' it in latent space towards a desired distribution, which is a  challenging problem~\citep{jahanian:iclr20}.
 
%Algorithms that co-author content \emph{with} human designers need to interact with human specified preferences. For example, generative adversarial networks~\citep{goodfellow:nips14, karras:iclr2018} generate realistic looking images after training on a large dataset of human specified examples.

%However, the controls for exploring the space of generated images is the latent vector, which a human designer finds opaque~\citep{schrum:gecco20}. Relying on the human designer to sift through randomly generated examples from a GAN is expensive. At the same time, GANs inherit the biases of the training distribution, and ``steering'' them in latent space towards a desired output distribution is a challenging problem~\citep{jahanian:iclr20}.

%Algorithms that co-author content \emph{with} human designers need to interact with human specified preferences. For example, generative adversarial networks~\citep{goodfellow:nips14, karras:iclr2018} generate realistic looking images after training on a large dataset of human specified examples. However, the controls for exploring the space of generated images is the latent vector, which a human designer finds opaque~\citep{schrum:gecco20}. Relying on the human designer to sift through randomly generated examples from a GAN is expensive. At the same time, GANs inherit the biases of the training distribution, and ``steering'' them in latent space towards a desired output distribution is a challenging problem~\citep{jahanian:iclr20}.

When desired attributes can be formulated as an objective, one approach is to explore the latent space using derivative-free optimization algorithms such as \hbox{CMA-ES}~\citep{hansen:cma16}. \citet{bontrager:icbtas18} named this approach latent variable evolution (LVE). Later, \citet{volz:gecco18} proposed using GANs to automatically author Mario levels and demonstrated how LVE can extract level scenes with specific attributes from latent space.

%When desired attributes can be formulated as an objective, latent space can be explored using derivative-free optimization algorithms such as CMA-ES~\citep{hansen:cma16}. \citet{bontrager:icbtas18} named this approach latent variable evolution (LVE). Later, \citet{volz:gecco18} proposed using GANs to author Mario levels and demonstrated how LVE can extract levels with specific attributes from latent space.

The LVE approach is limited to attributes that are easily specifiable as an objective. A human designer may not know \emph{exactly} what kind of content they want, but instead have some intuition on how they would vary content when exploring GAN generated levels. For example, the designer may want to have levels that are of varying difficulty; while it is hard to specify difficulty as an objective, a designer can choose from automatically generated levels of different number of enemies or obstacles. 

%The LVE approach is limited to content that is easily specifiable as an objective. A human designer may not know exactly what kind of content they want. However, they often have intuitions on how they would vary content when exploring GAN generated levels. For example, the designer may want to explore levels with different numbers of enemies or obstacles. However, the human designer may still want to optimize for aspects like playability or specific constraints.

\begin{figure}[t!]
\centering
\includegraphics[width=1.0\linewidth]{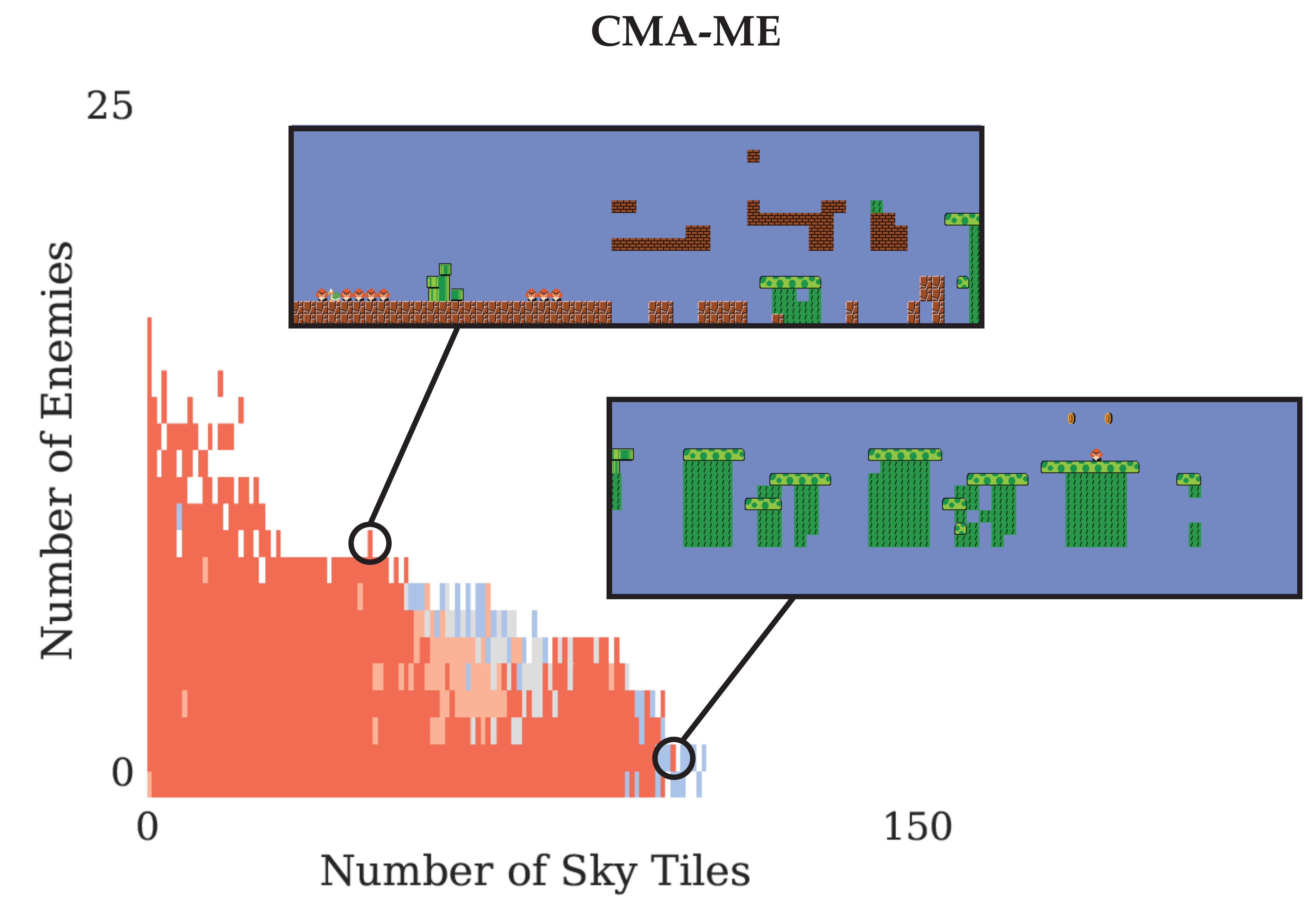}
\caption{Mario scenes returned by the CMA-ME quality diversity algorithm, as they cover the designer-specified space of two level mechanics: number of enemies and number of tiles above a given height. The color shows the percentage of the level completed by an A* agent, with red indicating full completion.}
\label{fig:cmame_diversity}
\end{figure}

We call the above problem \emph{latent space illumination} (LSI). Formally, given an objective function and additional functions which measure different aspects of gameplay, we want to extract a collection of game scenes that collectively satisfy all output combinations of the gameplay measures. For each output combination, the representative scene should maximize the objective function. 

\textit{Quality diversity} (QD) algorithms~\citep{pugh:frontiers16} are a class of algorithms designed to discover a diverse range of high-quality solutions  with several specialized variants designed to explore continuous search spaces.

Our goal in this paper is twofold: First, we wish to find out whether QD algorithms are effective in illuminating the latent space of a GAN, in order to generate high-quality level scenes with a diverse range of desired level characteristics, while still maintaining stylistic similarity to human-authored examples. Second, we want to compare the state-of-the-art QD algorithms in this domain and provide quantitative and qualitative results that illustrate their performance.

A large-scale experiment shows that the QD algorithms MAP-Elites, MAP-Elites (line) and CMA-ME significantly outperform CMA-ES and random search in finding a diverse range of high-quality scenes.\footnote{The source code of the algorithms is available at \url{https://github.com/icaros-usc/MarioGAN-LSI}.} Additionally, \mbox{CMA-ME} outperformed the other tested algorithms in terms of diversity and quality of the returned scenes. We show generated scenes, which exhibit an exciting range of mechanics and aesthetics (Fig.~\ref{fig:cmame_diversity}).  A user study shows that the diverse range of level mechanics translates to different subjective ratings of each scenes' difficulty and appearance, highlighting the promise of  quality diversity algorithms in generating diverse, high-quality content by searching the latent space of generative adversarial networks.

\section{Background}
%\subsection{Procedural Content Generatio
\noindent\textbf{Procedural Content Generation.}
Procedural content generation (PCG) refers to
%the body of work focused on 
creating game content algorithmically, with limited human input~\citep{shaker:book16}. Game content can be any asset (e.g., game mechanics, rules, dialog, models, etc) required to realize the game for its players. Pioneering work in PCG dates back to the 1980s to address memory limitations for storing large video game levels on computers. The growing interest in realistic graphics in the 1990's necessitated the development of procedural modelling algorithms~\citep{smelik:cgf2014} to generate complex models such as trees and terrain to ease the burden on graphic artists. Much PCG research in both industry and academia has focused on generating playable levels. In general, the problem of generating content that fulfils certain constraints can be approached by evolutionary solutions
~\citep{togelius2011search} or constraint satisfaction methods~\citep{smith2011answer}. An emerging area of research is PCG via machine learning (PCGML) which aims to leverage recent advancements in machine learning (ML) to generate new content by treating existing human authored content as training data~\citep{summerville2018procedural}. Previous work in PCGML has enabled automatic generation of video game levels for the Super Mario Bros. using LSTMs~\cite{summerville2016super}, Markov Chains~\cite{snodgrass2014experiments} and probabilistic graphical models~\cite{guzdial2016game}.  

%Before Amy Edits: 
    Two recent advancements in PCGML are studies by \citet{volz:gecco18} and \citet{giacomello:gem18} who independently demonstrated the successful application of generative adversarial networks (GANs) to generate playable video game levels in an unsupervised way from existing video game level corpora. \citet{volz:gecco18} adapted the concept of latent variable evolution (LVE)~\citep{bontrager:icbtas18} to extract Mario scenes from the latent space of a GAN that targeted specific gameplay features. That work searched the latent space of the GAN utilizing the popular Covariance Matrix Adaptation Evolution Strategy (\mbox{CMA-ES})~\citep{hansen:ec01} for latent variable inputs that would make the GAN produce level scenes with desired properties. Scenes with targeted gameplay features were obtained through carefully crafted single-objective functions, named fitness functions, that carefully balanced weighted distance from desired gameplay properties on the generated scenes.

\noindent\textbf{Quality Diversity.} While the approach employed by \citet{volz:gecco18} demonstrated a promising synergy between generative models and evolutionary computation for PCG, other works in PCG displayed the potential of quality diversity (QD) to generate meaningfully diverse video game content~\citep{gravina2019procedural}. Unlike traditional optimization methods, QD algorithms aim to generate high quality solutions that differ across specified attributes. Consider the example of generating Mario levels with specific properties. Instead of incorporating the number of enemies or floor tiles into the fitness function, a QD algorithm can treat these measures as attributes. The QD algorithm still has the objective of finding solvable Mario levels, but must find levels that contain all combinations of attributes (number of enemies, percentage of floor coverage). \citet{mouret2015illuminating} coined the term \textit{illumination algorithms} for quality diversity (QD) algorithms that create an organized mapping between solutions and their associated attributes, which are called behavioral characteristics (BCs). After the QD algorithm generates an organized palette of scenes, stitching algorithms can combine several scenes together to form a cohesive level~\citep{green2020mario}.

Developed concurrently with our approach is CPPN2GAN~\citep{schrum:cppn:geeco20}, which generates full levels for both Super Mario Bros and Zelda. The paper proposes optimizing the latent space of a GAN  with a special type of encoding, a compositional pattern producing network  (CPPN, \cite{stanley:gpem07}), which captures patterns with regularities. The paper introduces a type of latent space illumination with a vanilla version of the quality diversity algorithm MAP-Elites \citep{mouret2015illuminating}, described in the next section. It focuses on simultaneously searching several latent vectors at once to generate a full level created by ``stiching'' together GAN-generated scenes. Instead, our focus is on assessing the performance of QD algorithms in generating a variety of scenes with desired characteristics, and in measuring modern MAP-Elites variants that excel at the exploration of continuous domains. Our work is also related with conditional generative models~\cite{hald2020procedural,snodgrass2014experiments,ping2020conditional}. While it is possible to condition GANs on desired BCs, there is no guarantee that the generated scenes will have the properties specified by the conditioning input. Additionally, conditional generative models require retraining for each new set of BCs a human designer wishes to explore, where LSI can search the latent space of the same generative model without retraining.

\noindent\textbf{MAP-Elites.} \mbox{MAP-Elites} \citep{mouret2015illuminating} is a QD algorithm that searches along a set of explicitly defined attributes called behavior characteristics (BCs). These attributes collectively form a Cartesian space named the \emph{behavior space}, which is tessellated into uniformly spaced grid cells. MAP-Elites maintains the highest performing solution for each cell in behavior space (an \emph{elite}) with the product of the algorithm being a diverse archive of high performing solutions. The archive is initially populated with randomly sampled solutions. The algorithm then generates new solutions by selecting elites from the archive at random and perturbing each elite with small variations. The objective of the algorithm is both to expand the archive, maximizing the number of filled cells, and to maximize the quality of the elite within each cell. How the behavior space is tessellated is the focus of a variety of recent algorithms~\citep{smith:ppsn16,fontaine:gecco19}.

\noindent\textbf{MAP-Elites (line).} A common characteristic of many tasks is that high-performing solutions that exhibit diverse behaviors share significant similarities in their ``genotype'', that is in their search space parameters. Therefore,  \citet{vassiliades:gecco18} propose a variational operator, called “Iso+LineDD” which captures correlations between elites. When generating a new solution, in addition to applying a random variation to an existing elite, the operator adds a second random variation directed towards a second elite, essentially nudging the variation distribution towards other high performing solutions. We denote MAP-Elites with this operator ME (line).

\begin{figure*}
\centering
\includegraphics[draft=false,width=0.45\textwidth]{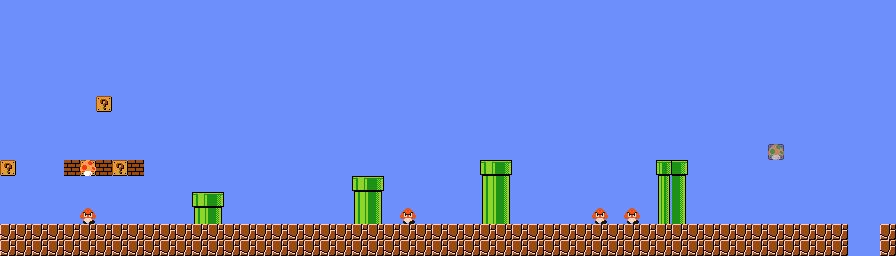}
\includegraphics[draft=false,width=0.45\textwidth]{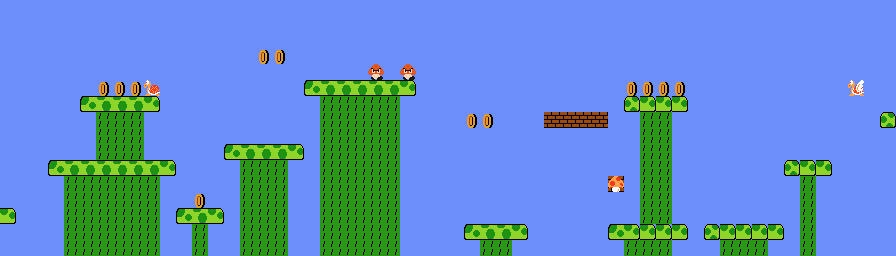}
\caption{Ground truth scenes 1 (left) and 2 (right) for KL-divergence metric.}
\label{fig:KL-groundtruth}
 \end{figure*}

\noindent\textbf{CMA-ES.}\label{subsec:CMA-ES} The Covariance Matrix Adaptation Evolution Strategy (\mbox{CMA-ES}) is a second-order derivative-free optimizer for single-objective optimization of continuous spaces~\citep{hansen:cma16}. The algorithm belongs to a family of algorithms named evolution strategies (ES), which specialize in optimizing continuous spaces by sampling a population of solutions, called a generation of solutions, and gradually moving the population towards areas of highest fitness. CMA-ES models the sampling distribution of the population as a multivariate normal distribution. The algorithm adjusts its sampling distribution by ranking solutions based on their fitness and estimating a new covariance matrix that maximizes the likelihood of future successful search steps.

%The main mechanisms steering CMA-ES are the selection and ranking of the $\mu$ fittest solutions, which update the next generation's next sampling distribution, $\mathcal{N}(m, C)$. CMA-ES maintains a history of aggregate changes to $m$ called an evolution path, which provides benefits to search that are similar to momentum in stochastic gradient descent. 

%Evolution strategies (ES) are a family of evolutionary algorithms that specialize in optimizing continuous spaces by sampling a population of solutions, called a generation, and gradually moving the population toward areas of highest fitness. One canonical type of ES is the  $(\mu / \mu, \lambda)$-ES, where a population of $\lambda$ sample solutions is generated, then the fittest $\mu$ solutions are selected to generate new samples in the next generation. The $(\mu / \mu, \lambda)$-ES recombines the $\mu$ best samples through a weighted average into one mean that represents the center of the population distribution of the next generation.

%The Covariance Matrix Adaptation Evolution Strategy (CMA-ES) is a particular type of this canonical ES, which is one of the most competitive derivative-free optimizers for single-objective optimization of continuous spaces~\citep{hansen:gecco10}.

\noindent\textbf{CMA-ME.} \label{subsec:CMA-ME}The Covariance Matrix Adaptation MAP-Elites (\mbox{CMA-ME})~\citep{fontaine:gecco20}
is a recent hybrid algorithm which incorporates CMA-ES into MAP-Elites. The algorithm improves the efficiency in which new archive cells are discovered and the overall quality of elites within the archive. CMA-ME maintains a number of individual CMA-ES-like instances, named \textit{emitters}. We use a specific type of emitter named \textit{improvement} emitter, which was shown to outperform MAP-Elites and ME (line) in the  strategic card game Hearthstone~\cite{fontaine:gecco20}. Improvement emitters rank solutions by prioritizing those that fill previously undiscovered cells in the archive. 
Solutions that belong to existing cells in the map are subsequently ranked based on the improvement in fitness over existing cells.  This enables improvement emitters to dynamically adapt their goals based on feedback from how the archive changes.

\section{Mario Scene Evaluation}
We used the Mario AI Framework\footnote{https://github.com/amidos2006/Mario-AI-Framework} to evaluate each of the generated scenes. We evaluate each scene by treating it as a playable level; actual levels are longer and can be generated by ``stiching'' together multiple scenes~\citep{green2020mario}. 

Following previous work~\citep{volz:gecco18,awiszus2020toad}, we approximate playability of a scene by how far through the scene A* reaches; Specifically, we define as ``fitness'' of a scene the amount of progress by an AI agent playing the scene (percentage of completion in the horizontal direction). We use the A* agent that won the 2009 Mario competition.\footnote{https://www.youtube.com/watch?v=DlkMs4ZHHr8} We additionally define three different types of behavioral characteristics (BCs), which allow for a diverse set of level mechanics.\footnote{One could also combine the BCs from the three different types, e.g., have an archive with KL-divergence and number of enemies.}

\noindent\textbf{Representation-Based.} We define a set of BCs that capture stylistic aspects of the Mario scene's representation, based on the distribution of tiles. These BCs do not depend on the agent's playthrough:
\begin{enumerate}
    \item Sky tiles: These are game objects, e.g., blocks, question blocks, coins, that are above a certain height value. A large number implies that  there are many game elements above ground, and the player would need to jump to higher tiles. 
    \item Number of enemies: A larger number of enemies generally results in higher difficulty and requires the player to perform more jumps to navigate throughout the scene.
\end{enumerate}

\noindent\textbf{Agent-Based.} We incorporate the agent-based BCs of previous work~\citep{khalifa:gecco18}, which are computed after one playthrough by the agent. The BCs are binary, representing whether the playthrough satisfied a given condition. This results in an 8-dimensional BC-space. The 8 conditions are: (1) performing a jump, (2) performing a high jump (height of jump is above a certain threshold), (3) performing a long jump (horizontal distance is above a certain threshold), (4) stomping on an enemy,  (5) killing an enemy using a koopa shell, (6) having an enemy die because of falling out of the scene, (7) collecting a mushroom, and (8) collecting a coin. 
% \begin{table}[h!]
%     \centering
%     \begin{tabular}{p{2cm}|p{5.5cm}}
%         BC & Description \\\hline
%         Jump & True if the player jumped \\
%         High Jump & True if the player jumped above a certain threshold \\
%         Long Jump & True if the player’s horizontal traversed distance from a jump is larger than a certain value\\
%         Stomp & True if the player stomped on an enemy \\
%         Shell Kill & True if the player killed an enemy using a koopa shell  \\
%         Fall Kill & True if an enemy dies because of falling out of the scene\\
%         Mushroom &  True if the player collected a mushroom\\
%         Coin & True if the player collected a coin\\
%     \end{tabular}
%     \caption{Agent-Based Behavioral Characteristics from~\citet{khalifa:gecco18}.}
%     \label{tab:8Binary}
% \end{table}

\noindent\textbf{KL-Divergence.} A common goal in procedural content generation is to generate scenes  with different degrees of stylistic similarity to human-designed examples. We use the tile pattern Kullback–Leibler divergence metric~\citep{lucas2019tile} to measure the structural similarity between two Mario scenes. We picked two stylistically different human-designed scenes from the Mario AI Framework, shown in Fig.~\ref{fig:KL-groundtruth}, and we set the behavior characteristics to be the tile pattern KL-divergence between the ground truth scene and generated scene, resulting in a 2-dimensional BC space. 

%In Eq.~\ref{eq:fitness}, $d_{completed}$ is the distance in
%BC Description and VisualizationI.  

%LEVELEVALUATIONWe  used  Mario  AI  Framework  to  evaluate  each  level  gen-erated  by  the  GAN.  According  to  the  AI  agent’s  result,  weassigned  a  fitness  valuefto  each  level.  We  definedfto  bethe percentage of completion, as Equation   1 shows

\section{Experiments}

Our experiments compare the performance of random search, CMA-ES, MAP-Elites, MAP-Elites (line) and CMA-ME on the problem of latent space illumination.

We ran each of the 5 algorithms for 20 trials, 10,000 evaluations each, for each of the three different BC combinations. This resulted in a total of 300 trials. We ran all trials in parallel in a university cluster with multiple nodes running on dual Intel Xeon L5520 processors. Each trial lasted approximately 7 hours. 

\noindent\textbf{GAN Model.} We use a deep convolutional GAN (DCGAN) as in the study by \citet{volz:gecco18}, trained with the WGAN algorithm~\citep{martin2017wasserstein}. Following their implementation, we encode the training levels by representing each of the 17 different tile types by a distinct integer, which is converted to an one-hot encoded vector, before passed as input to the discriminator. We pad each training level to a  $64\times 64$ matrix, and since there are 17 channels, one for each possible tile type, each input scene to the discriminator is $17 \times 64 \times 64$. For the generator, we set the size of the latent vector to be 32, resulting in a 32-dimensional continuous search space. We refer the reader to the study by~\citet{volz:gecco18} for the details of the architecture. 

We train the DCGAN with RMSprop for 5000 iterations, a learning rate of $5e^{-5}$ and a batch size of 32. The discriminator iterates 5 times before the generator iterates once. We used for training 15 original levels from the Mario AI competition framework.\footnote{\url{https://github.com/amidos2006/Mario-AI-Framework/tree/master/levels/original}} Fig.~\ref{fig:KL-groundtruth} shows scenes from two levels of the training data. 

To evaluate the different search algorithms, we input the latent vector of size 32 to the generator, and we crop the $17 \times 64 \times 64$ output to a $17 \times 16 \times 56$ playable level for evaluation.

%The only difference between our GAN and proposed GAN by~\citet{volz:gecco18} is the number of input channels. In our approach, we have 17 different channels while the original GAN have only 10 different channels. The increase is due to having more variants of enemies (5 different enemies), different type of solid tiles (3 different types), two different types of question mark blocks, and floating coins. On the other hand, we collapsed all the tube pieces into one channel and let the visualizer handle the tube look (as long as we have two tiles beside each other, the tube will look correctly).

%We refer the reader to~\cite{volz:gecco18} for the architecture of the network and details about the encoding of the levels.

\noindent\textbf{Search Parameters and Tuning.} We tuned each algorithm based on how well it covered the representation-based behavior space and we then used the same parameters for all three behavioral characteristics. We set population size $\lambda = 17$ and mutation power $\sigma = 0.5$ for CMA-ES. A single run of CMA-ME deploys 5 improvement emitters with $\lambda = 37$. We set the mutation power for CMA-ME and \hbox{MAP-Elites} $\sigma = 0.2$. For ME (line), we set the isotropic mutation $\sigma_1 = 0.02$ and the mutation for the directional distribution $\sigma_2 = 0.2$. The initial population for MAP-Elites and ME (line) was 100. 

In random search, we generate solutions  by sampling directly the GAN's latent space from the same distribution that we used to train the generator network: a normal distribution with zero mean and variance equal to 1. We used the same method to generate solutions for the initial population of MAP-Elites and ME (line).

\noindent\textbf{Map Sizes.} We performed an initial run of the experiment and we observed the maximum and minimum of values of the behavioral characteristics covered by each algorithm. This provided a rough estimate of the range of each BC.

For the representation-based BCs, we set the range of sky tiles to [0,150] and the number of enemies to [0,25]. The map size was $151 \times 26$, where each cell corresponded to an integer value of the BC. The eight agent-based binary BCs form an eight-dimensional map of $2^8 = 256$ cells. Finally, we set the KL-divergence ranges to [0, 4.5] for both groundtruth levels, and the resolution of the map was $60\times60$.

\noindent\textbf{Metrics.} We evaluate all five algorithms, random search, CMA-ES, MAP-Elites, ME (line) and CMA-ME, with respect to the diversity and quality of solutions returned. For comparison purposes, we assign the solutions by CMA-ES and random search to a grid location on what their BC would have been and populate a pseudo-archive.

\noindent\textit{Percentage of valid cells:} This is the percentage of scenes in the archive returned by the algorithm that are completed from start to end by the A* agent, which is equivalent to having a fitness of 1.0. This is an indication of the quality of the solutions found.

\noindent\textit{Coverage:} This is the percentage of cells in the archive produced by an algorithm, computed as the number of cells divided by the total map size. The measure indicates how much of the behavior space is covered. 

\noindent\textit{QD-Score:} The QD-Score metric was proposed by~\citet{pugh2015confronting} as the sum of fitness values of all elites in the archive and has become a standard QD performance measure. The measure distills both the diversity and quality of elites in the archive into a single value.

%We used $\sigma = 0.2$ for Map-Elites. We selected the same mutation power for CMA-ME, while we set 

%population size $\lambda = 14$ and $\sigma = 0.5$ for CMA-ES,
\section{Results}

\begin{table*}[t]
\centering
\resizebox{1.0\linewidth}{!}{
\begin{tabular}{l|cccc|cccc|cccc}
\hline
             & \multicolumn{4}{c|}{Representation-Based BCs}          & \multicolumn{4}{c|}{Agent-Based BCs} & \multicolumn{4}{c}{KL-Divergence}  \ \\ 
    \toprule
Algorithm  & Valid / All   & Coverage & Valid / Found & QD-Score   & Valid / All & Coverage & Valid / Found & QD-Score   & Valid / All &  Coverage & Valid / Found&  QD-Score\\
    \midrule
Random &8.35\% & 11.1\% & 75.3\% & 385.1 &7.09\% & 8.9\% & 79.7\%& 20.2 &5.10\%   & 12.5\% & 40.8\% & 331.5\\
CMA-ES  &7.44\% & 8.0\% & 93.0\% & 308.6  &7.43\% &8.3\% & 89.6\% &  19.8 &4.11\% & 7.5\% & 54.8\%& 210.6 \\
ME    &15.15\% & 19.4\% &78.1\% &  692.5  &7.66\% & 8.8\% & 87.0\% &    20.4 &9.98\% & 15.5\% & 64.4\% &    485.6\\
ME (line)  &15.31\%   & 18.9\% & 81.0\%  & 682.7  &7.06\%& 8.2\% & 86.1\% &  18.9  &10.18\%& 15.4\% & 66.1\% & 488.0\\
CMA-ME & 16.35\% & 21.5\% & 76.1\% & 776.8 &7.90\% & 9.4\% & 84.0\% & 21.6 &11.08\%  & 17.4\% & 63.7\% &  551.3\\
  \bottomrule
\end{tabular}
}
\caption{Results: Average percentage of cells with fitness 1.0 (Valid / All), percentage of cells found (Coverage), percentage of cells found with fitness 1.0 (Valid / Found), and QD-score after 10,000 evaluations. }
\label{tab:results}
\end{table*}

\begin{figure*}[t!]
    \centering
    \begin{subfigure}[t]{0.3\textwidth}
        \centering
        \includegraphics[width=\textwidth]{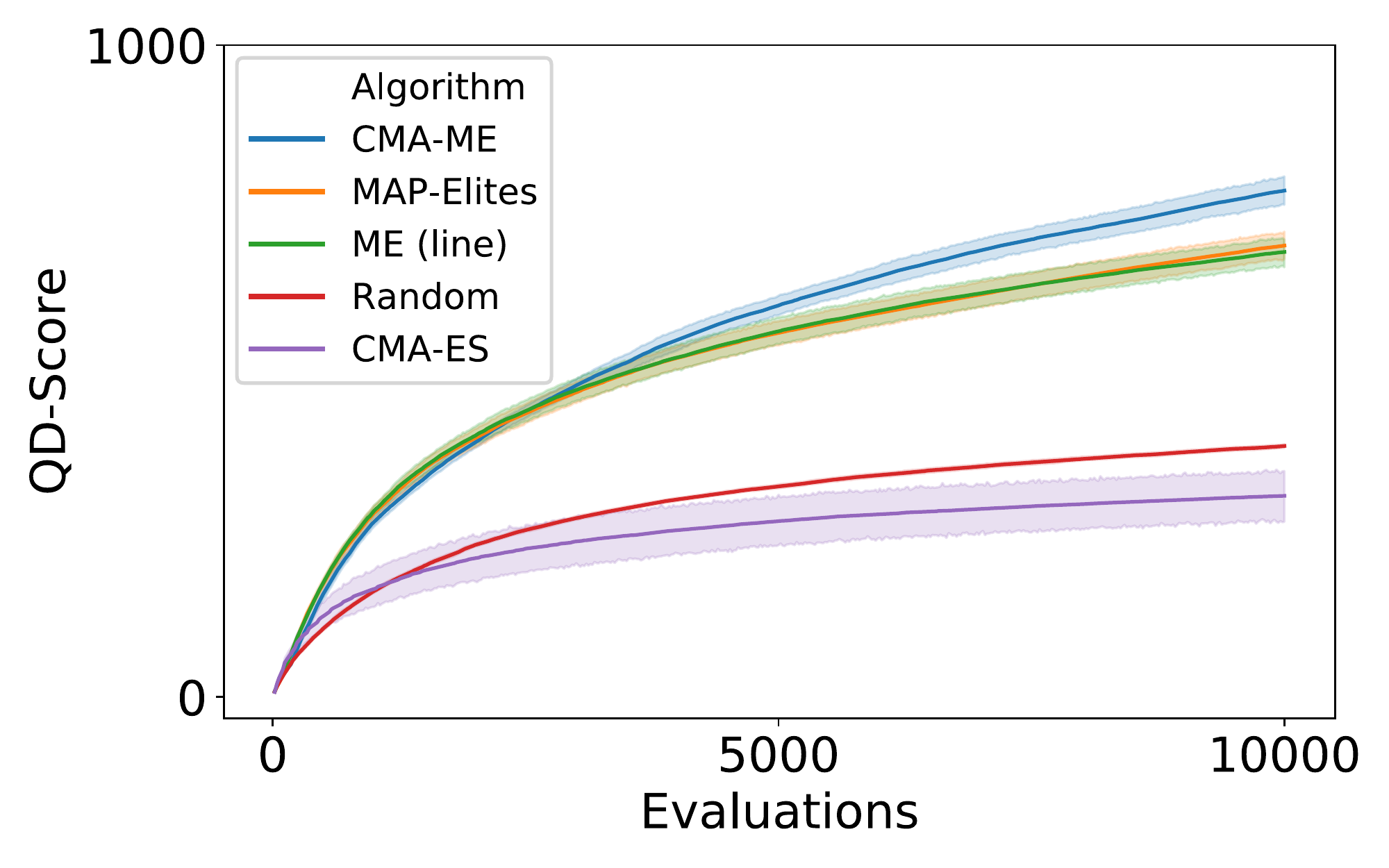}
        \label{fig:representation-BC}
        \caption{Representation-based BCs}
    \end{subfigure}
    \begin{subfigure}[t]{0.3\textwidth}
        \centering
        \includegraphics[width=\textwidth]{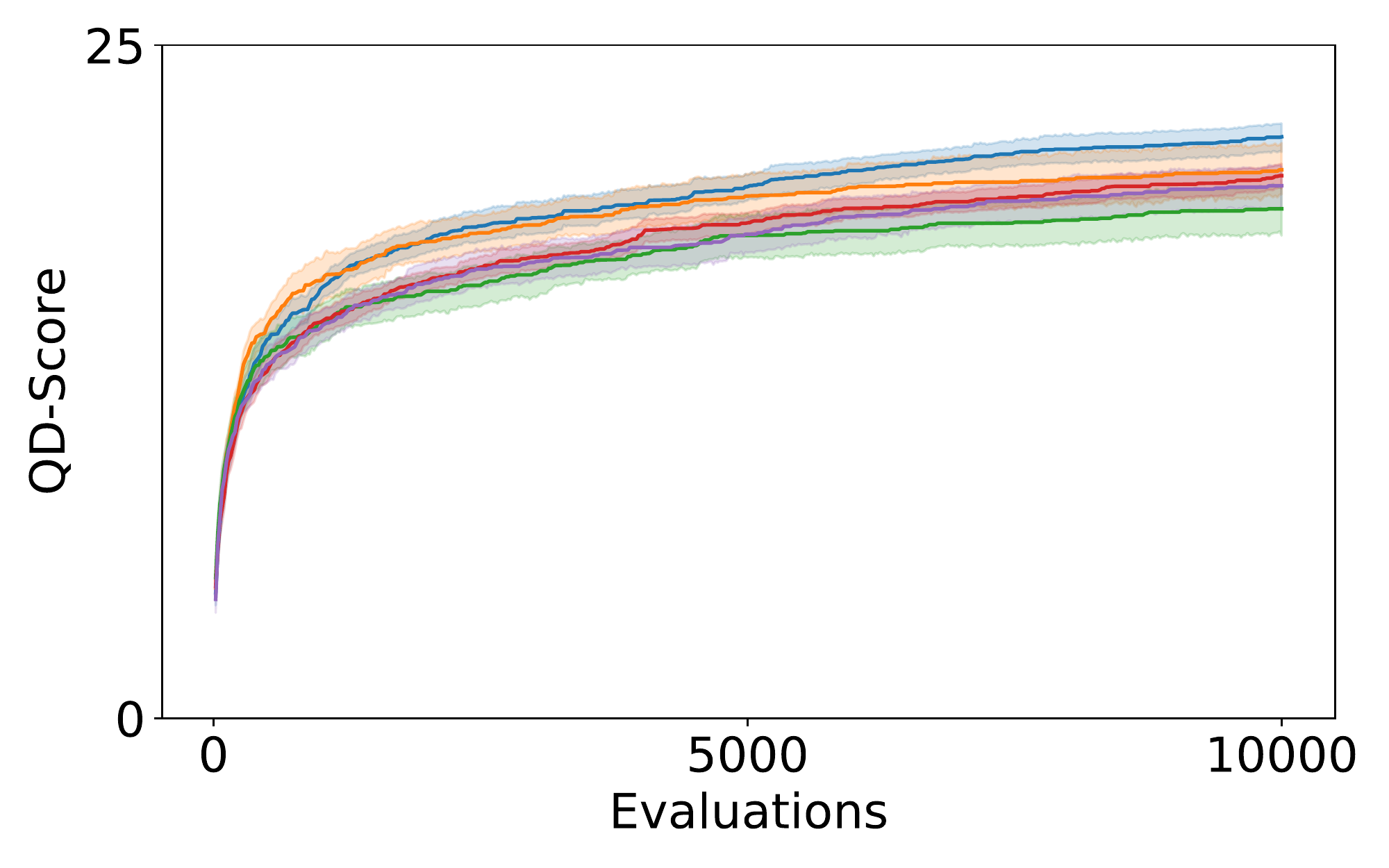}
        \label{fig:agent-based-BC}
        \caption{Agent-based BCs}
    \end{subfigure}
    \begin{subfigure}[t]{0.3\textwidth}
        \centering
        \includegraphics[width=\textwidth]{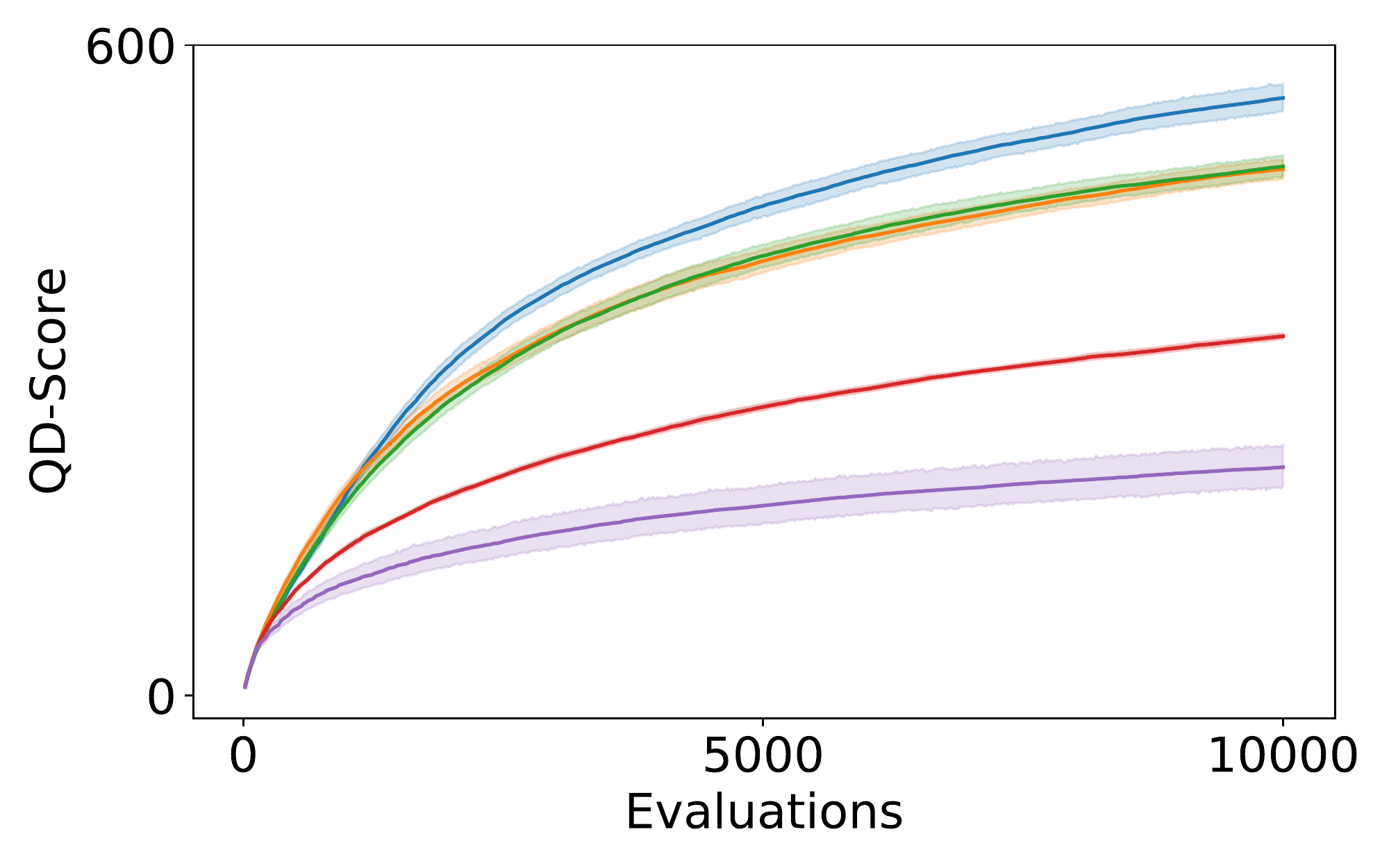}
        \label{fig:KL}
        \caption{KL-divergence}
    \end{subfigure}
% \subfigure[Representation-Based BCs]{\centering\includegraphics[draft=false,width=0.3\textwidth]{figs/MarioGAN.pdf}\label{fig:representation-BC}}
% \subfigure[Distribution of Elites]
% {
% \includegraphics[draft=false,width=0.26\textwidth]{figs/MarioGAN.pdf}
% }
% \subfigure[Agent-Based BCs]{\centering\includegraphics[draft=false,width=0.3\textwidth]{figs/8Binary.pdf}\label{fig:agent-based-BC}}
% \subfigure[KL-Divergence]{\centering\label{fig:KL}\includegraphics[draft=false,width=0.3\textwidth]{figs/KL.pdf}}
    \caption{QD-Scores over time for each behavioral characteristic. }
    \label{fig:QD-compare}
\end{figure*}

\begin{figure*}[t!]
\vspace{-1em}
    \centering
    \hspace{1.2em}
    \begin{subfigure}[t]{.3\textwidth}
        \centering
        \includegraphics[width=\textwidth]{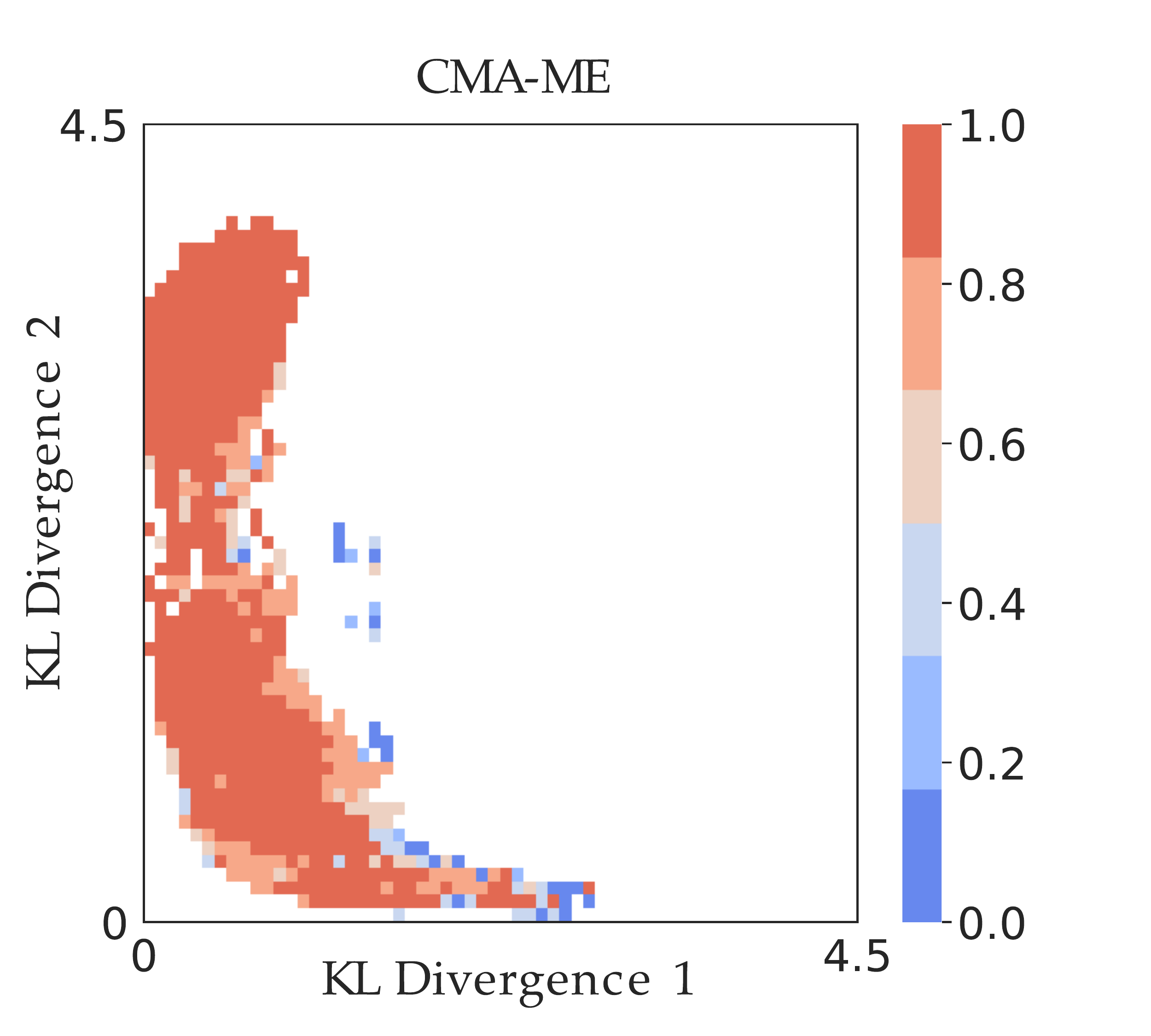}
    \end{subfigure}
     %   \hspace{0.1em}
    \begin{subfigure}[t]{.3\textwidth}
        \centering
        \includegraphics[width=\textwidth]{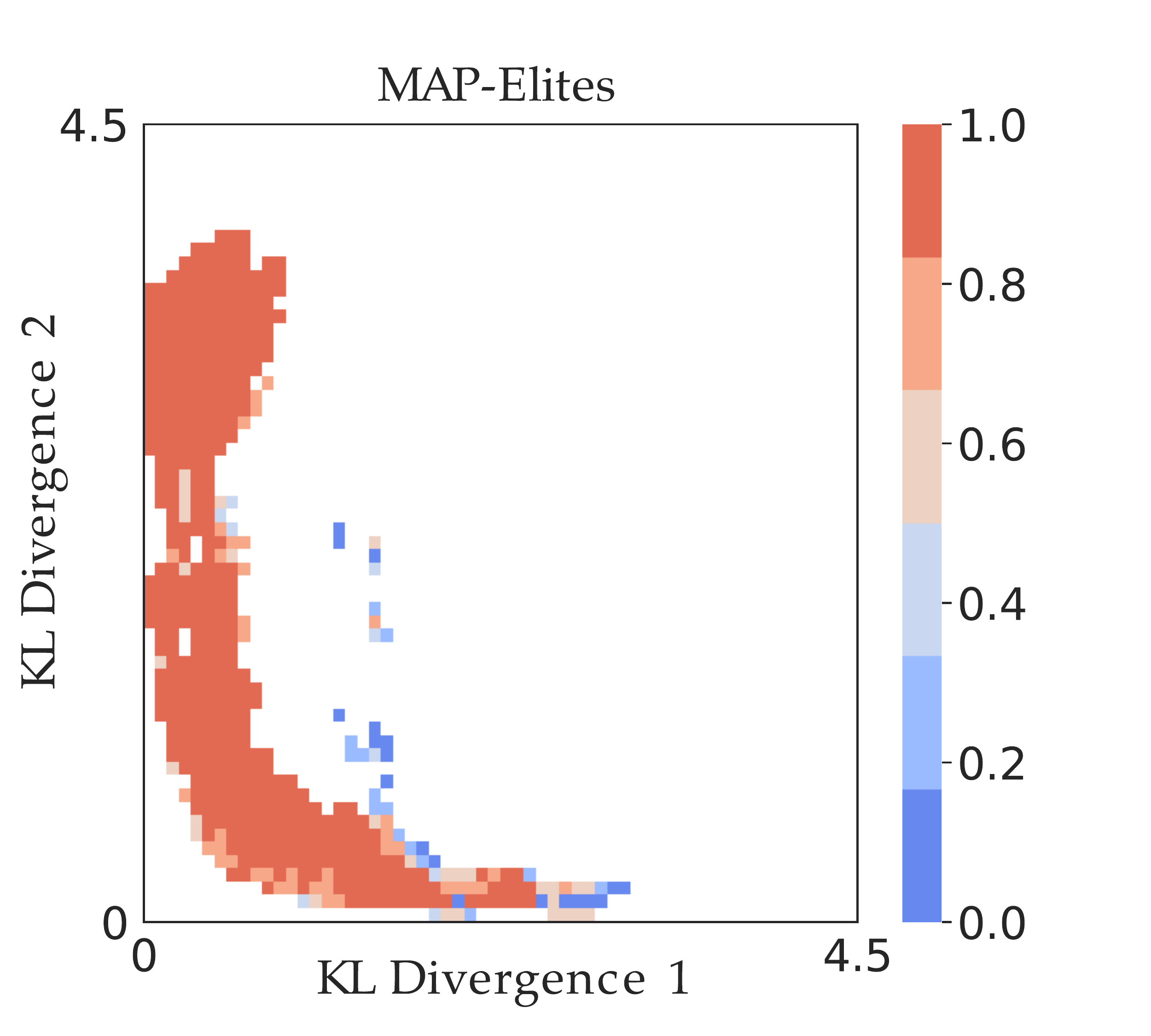}
    \end{subfigure}
      %  \hspace{0.1em}
    \begin{subfigure}[t]{.3\textwidth}
        \centering
        \includegraphics[width=\textwidth]{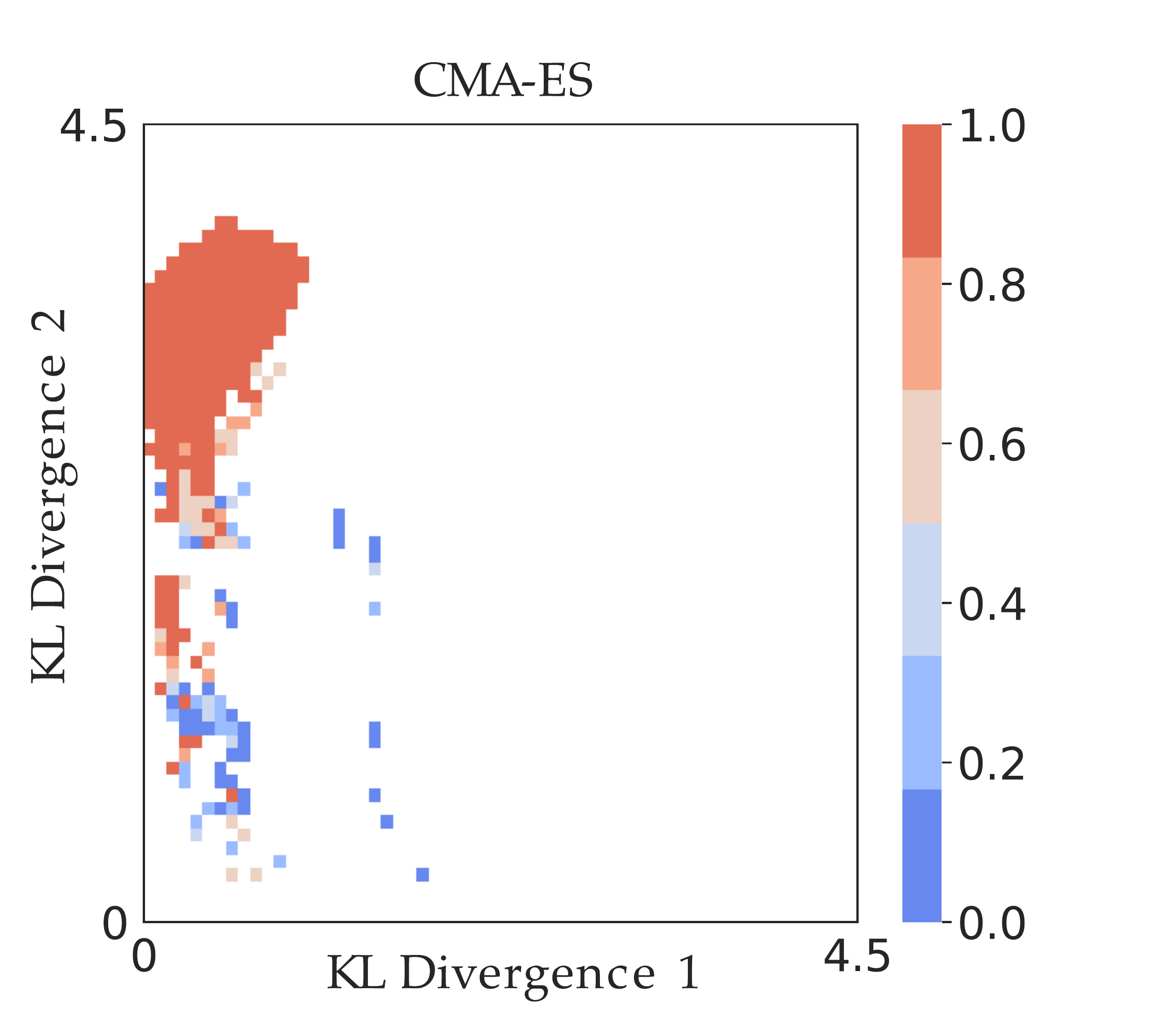}
    \end{subfigure}
\caption{Archive for the KL-divergence behavioral characteristic metric.}
\label{fig:map-elites-KL}
\end{figure*}

% \begin{figure*}[t!]
% \includegraphics[draft=false,width=1.0\textwidth]{figs/lvl-3.png}
% \caption{Training level 2}
% \end{figure*}

\noindent\textbf{Performance.} Table~\ref{tab:results} summarizes the performance of each algorithm. Fig.~\ref{fig:QD-compare} shows improvement in QD-score over evaluations for each algorithm, with $95\%$ confidence intervals.

First, we observe that all QD algorithms, i.e., MAP-Elites, ME (line) and CMA-ME outperform CMA-ES and random search in the representation-based and KL-divergence BCs. This is expected, since CMA-ES optimizes only for one objective, the playability of the scenes, rather than exploring a diverge range of level behaviors. Random search works poorly; the reason is that we sample from the same distribution that we used for training the GAN, thus the generated solutions follow the tile distribution of the training data, which covers only a small portion of the behavior space.

%low-dimensional behavior spaces are distorted~\citep{fontaine:gecco20}: when mapping a high-dimensional sampling uniformly from a high dimensional search space covers only a small region of the distribution of solutions in behavior space.
%Random search also returns the smallest percentage of valid solutions. This is expected since finding a high quality solution by  chance  is very  unlikely  for large  search  spaces.

Second, CMA-ME outperforms the other QD algorithms in the representation-based and KL-divergence BCs. This matches previous work~\citep{fontaine:gecco20}, where \mbox{CMA-ME} outperformed these algorithms in the Hearthstone strategic game domain. We attribute this to the fact that \mbox{CMA-ME} benefits by sampling from a dynamically changing Gaussian (as in \mbox{CMA-ES}) rather than a fixed distribution shape. Fig.~\ref{fig:map-elites-KL} shows three example archives of elites for \mbox{CMA-ME}, \mbox{MAP-Elites} and \mbox{CMA-ES}, illustrating the ability of \mbox{CMA-ME} to cover larger areas of the map.

We observe that ME (line) performs similarly to \mbox{MAP-Elites}. \mbox{ME (line)} relies on the assumption that different elites in the archive have similar search space parameters. We estimated the similarity of the elite hypervolume as defined in \citet{vassiliades:gecco18}, and found low mean values for the representation-based (0.60) and the KL-divergence (0.58) maps, which explains the lack of improvement from the operator in this domain. 

%It is likely that the latent parameters of the GAN that form levels of very different tile distributions do not match this assumption, and we will investigate this further in the future. 

%For instance, if we generate levels by searching through the latent space of a GAN, and we evaluate level quality by how playable they are, it is likely that the space of latent parameters of the GAN that form playable levels is concentrated in a small area of the parameter space, even though these levels may exhibit significant differences in how the tiles are distributed.

%This is likely not the case with the latent vector parameters that generate levels with very different tile distributions.

On the other hand, in the 8 binary agent-based BCs all algorithms perform similarly to random search. All of the algorithms performed poorly on these BCs, where each algorithm discovers less than 10\% of possible mechanic combinations. The main reason lies in the way the A* agent plays the levels; the agent is designed to reach the right edge of the screen as fast as possible, without caring much about its score. This forces the agent to avoid triggering gameplay mechanics. For example, in Fig.~\ref{fig:8Binary-lvls}(right) the agent rushes to the end without collecting the coins in the beginning of the level. The same holds for the training data; the human-authored levels covered only 20 out of the $2^8=256$ cells of the map, and there was no training level where the agent collected a mushroom or a coin. This makes the task of finding levels that trigger these BCs even more challenging.

\newcolumntype{M}[1]{>{\centering\arraybackslash}m{#1}}
\begin{figure*}[t!] 
  \begin{tabular}{M{15mm}M{74mm}M{74mm}}
    \centering

       Large Number of High Tiles & \includegraphics[draft = False, width=1.0\linewidth]{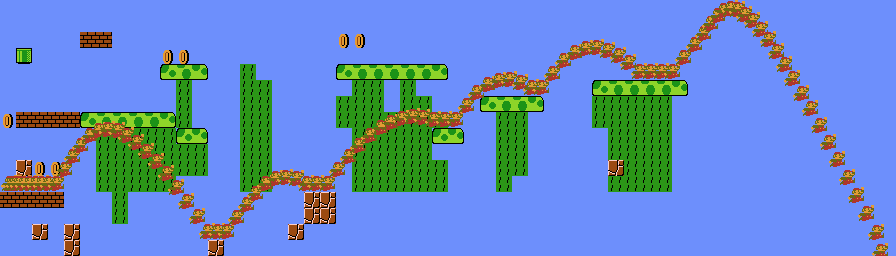} 
    %\caption{Initial condition} 
    \centering
    & 
    \includegraphics[draft = False, width=1.0\linewidth]{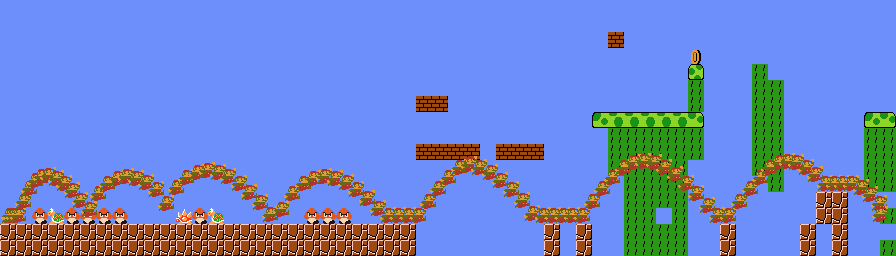}\\
   Small Number of High Tiles &
    \includegraphics[width=1.0\linewidth]{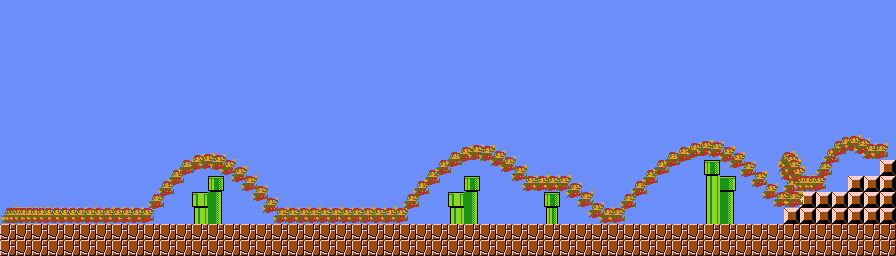} 
    %\caption{Initial condition} 
    &
    \includegraphics[width=1.0\linewidth]{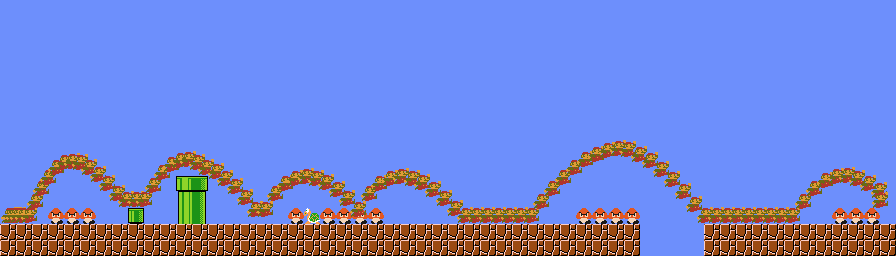}\\
    &  Small Number of Enemies  & Large Number of Enemies
\\
    \end{tabular}
    \caption{Generated scenes using CMA-ME for small and large values of sky tiles and number of enemies.} 
\label{fig:marioGANLevels}
\end{figure*}

\newcolumntype{L}[1]{>{\centering\arraybackslash}m{#1}}
\begin{figure*}[t!] 
  \begin{tabular}{M{15mm}M{74mm}M{74mm}}
    \centering
       \makecell{ } & \includegraphics[width=1.0\linewidth]{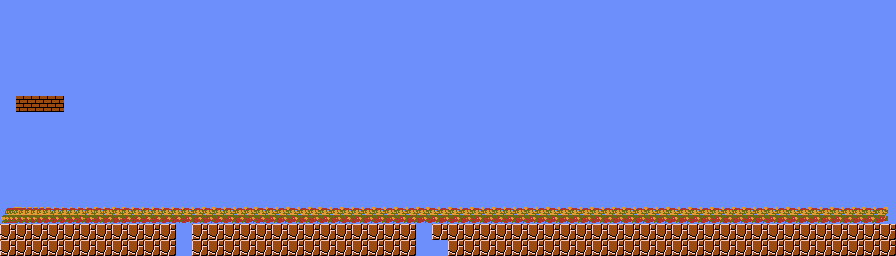} 
    & 
    \includegraphics[width=1.0\linewidth]{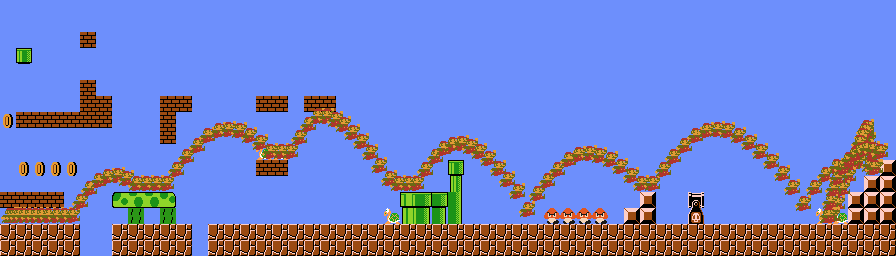} \\
    &  Min Number of Mechanics  & Max Number of Mechanics
    \\
    \end{tabular}
    \caption{Playable scenes with minimum (left) and maximum (right) sum value (6) of the 8 binary agent-based BCs.}
\label{fig:8Binary-lvls}
\end{figure*}

% \begin{figure*}[t!]
% \centering
% \includegraphics[draft=false,width=0.49\textwidth]{figs/min8Binary.png}
% \includegraphics[draft=false,width=0.49\textwidth]{figs/max8Binary.png}
% \caption{Playable levels with minimum (left) and maximum (right) sum value (6) of the 8 binary Agent-Based BCs.}
% \label{fig:8Binary-lvls}
% \end{figure*}

\newcolumntype{N}[1]{>{\centering\arraybackslash}m{#1}}
\begin{figure*}[t!] 
  \begin{tabular}{M{15mm}M{74mm}M{74mm}}
    \centering
       \makecell{Large KL-1} & \includegraphics[draft = False, width=1.0\linewidth]{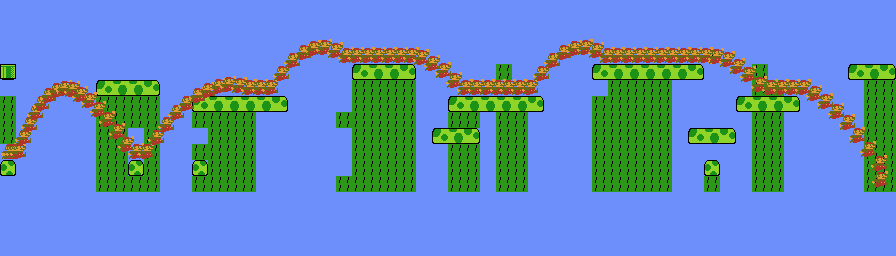} 
    %\caption{Initial condition} 
    \centering
    & 
    \includegraphics[draft = False, width=1.0\linewidth]{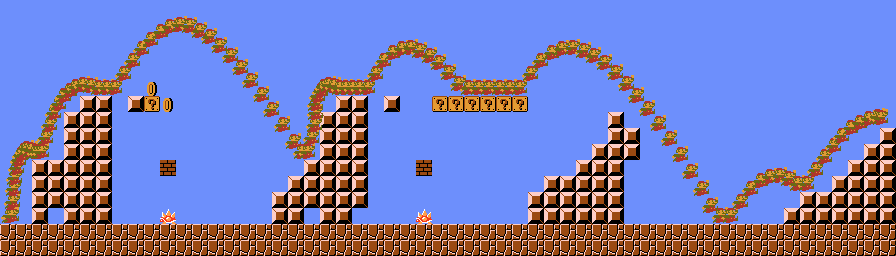} 
    %\caption{Initial condition} 
   \\
   \makecell{Small KL-1} &
    \includegraphics[draft = False, width=1.0\linewidth]{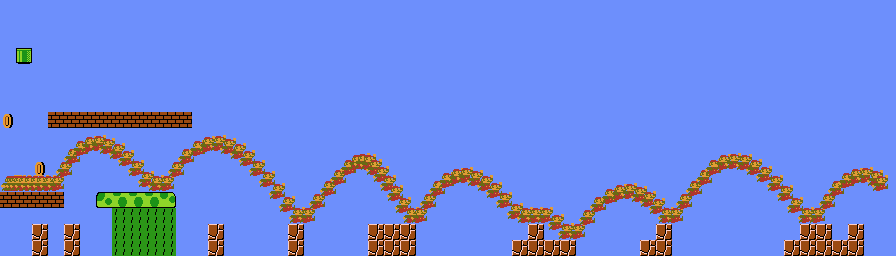} 
    %\caption{Initial condition} 
    &
    \includegraphics[draft = False, width=1.0\linewidth]{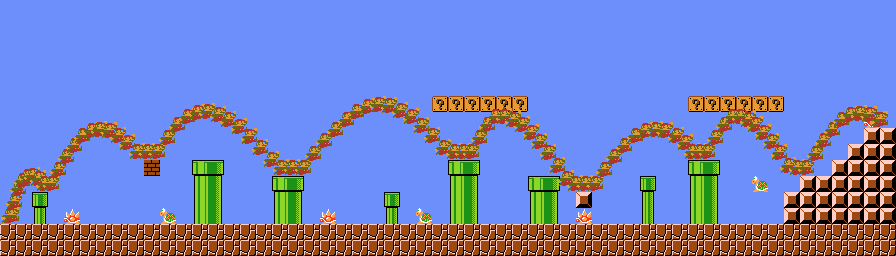} \\
    & Small KL-2 & Large KL-2
\\
    \end{tabular}
    \caption{Generated scenes using CMA-ME for small and large values of KL-divergence to each of the two groundtruth scenes.} 
\label{fig:KL-levels}
\end{figure*}

\noindent\textbf{Generated Levels.} We demonstrate generated levels by the CMA-ME algorithm that illustrate its ability to generate a diverse range of high-quality solutions. 
%We show example levels generated by the CMA-ME algorithm, that illustrate the diversity in the solutions found. Similar levels were also found  y the ME (line) algorithm. 

Figure~\ref{fig:marioGANLevels} shows four generated scenes from an archive generated by a single run of CMA-ME using the representation-based BCs. We selected the scenes from the map that had extreme values of the two BCs, the number of sky tiles and number of enemies. The scenes are significantly diverse, with the scene that maximizes each BC being filled with enemies and having multiple tiles above ground. Despite the large number of sky tiles at the level in the top-right, the agent finishes the scene without reaching most of them. This is a limitation of the representation-based BCs, which evaluate a scene based on the distribution of tiles and not on the agent's playthrough. 

We address the above limitation with  agent-based BCs. Fig.~\ref{fig:8Binary-lvls} shows two scenes generated by CMA-ME that minimize and maximize the sum of the agent-based BC values. The first scene has 0 value for all BCs and the agent simply runs a straight path towards the exit, while the second scene allows the agent to exhibit a variety of behaviors, including different types of jumps, stomping on an enemy and killing an enemy with a shell.

Finally, Fig.~\ref{fig:KL-levels} shows four scenes with small and large KL-divergence to each of the two groundtruth scenes in Fig.~\ref{fig:KL-groundtruth}. The scene that is stylistically similar to both groundtruths (bottom-left) combines ground tiles with gaps that force the agent to jump. The top left level maximizes divergence with the first groundtruth scene and minimizes divergence with the second; this results in the scene not having any ground tiles. Interestingly, the scene in the top right maximizes KL-divergence to both groundtruth scene by having tile types and enemies unseen in any of the groundtruth scenes.

\section{User Studies}
We have shown how to automatically generate levels that exhibit a diverse range of desired characteristics. Ultimately, the \textit{mechanical} diversity of the generated levels should translate to \textit{perceptual} diversity in how human players perceive the scenes. Our user study is motivated by \citet{sturtevant2020unexpected}, which demonstrates how mechanically similar levels can greatly vary in difficulty. Therefore, we conducted two user studies, where we asked users about their perception of scenes generated with the representation-based BCs and the KL-divergence BCs generated with CMA-ME.\footnote{Images of the selected scenes and videos of their playthrough by the AI agent are uploaded at: \url{https://icaros-usc.github.io/LSI-Mario-Level-Generation/}.}

%Beyond these objective metrics, we wish to address how these levels are perceived by actual users, who do not know the intrinsics of the generating algorithm. 

%Specifically, we expected that levels that have more sky tiles and enemies would be perceived as more challenging by users.\snnote{citations?} We also expected that levels with larger values of KL-divergence would be perceived as less similar to the corresponding groundtruths.

\noindent\textbf{Hypotheses.} 

\noindent\textbf{H1.} The perceived difficulty of the generated scenes increases with the number of sky tiles and enemies. 

\noindent\textbf{H2.} The perceived similarity of the generated scenes relative to the two groundtruth scenes decreases for larger values of KL-divergence. 

% \begin{figure}[t!]
% \centering
% \includegraphics[draft=false,width=0.5\columnwidth]{figs/difficulty.png}
% \caption{Mean ratings of perceived difficulty for the selected levels from the representation-based archive of Fig.~\ref{fig:cmame_diversity}. Different markers indicate different levels. The radius of the markers is proportional to the rating.}
% \label{fig:study-difficulty}
% \end{figure}

\noindent\textbf{Scene Difficulty.} We expect scenes that that have more sky tiles and larger number of enemies to be perceived as more challenging by human players. We picked 10 scenes uniformly from the representation-based archive, and presented to the users videos of the playthrough of an AI agent for each of the scenes in randomized order.

\textit{Dependent Measures.} We asked participants to rate how difficult it would be for a human player to complete the scene on a Likert scale from 1 (very easy) to 7 (very hard). At the end of the survey, we also asked them to briefly mention the factors affecting their rating.

\textit{Subject Allocation.} We recruited human participants through  Amazon’s Mechanical Turk service, and took several measures to ensure reliability of the results. All participants had approval rate of over 95\% and had completed more than 50 tasks.  We asked users a control question that tested their attention to the task, and eliminated data associated with a wrong answer, as well as incomplete data. We only considered users that had intermediate or higher experience of playing video games, resulting in 91 samples.

\textit{Analysis.} We fit a mixed-effects ordinal regression model to the data, with the number of sky tiles and number of enemies as fixed effects and the participant id as random effect. We normalized the numbers of sky tiles and enemies so that their range would be between 0 and 1. We found that both the number of sky tiles ($\beta = 4.35, t(817)= 11.56, p <0.001)$ and number of enemies  ($\beta = 0.94, t(817)= 2.48, p = 0.001)$ significantly predicted the perceived difficulty of the levels, supporting \textbf{H1}.  The $\beta$ values indicate that the number of sky tiles has a stronger effect on the difficulty of the level, compared to the number of enemies. This is because the AI agent attempted to complete the level as fast as possible and it ignored enemies most of the time. Indeed, most participants reported the frequency and length of jumps as the main factor affecting their rating.

%%Fig.~\ref{fig:study-difficulty} shows the selected levels in the map; the perceived difficulty of most of the selected levels increases with the number of sky tiles and enemies. 

\begin{figure}[t!]
\centering
%\begin{tabular}{cc}
%\begin{subfigure}[t!]{.5\columnwidth}
% \includegraphics[draft=false,width=1.0\textwidth]{figs/KL.png}
% \caption{}
% \label{fig:study-KL}
% \end{subfigure}& 
% \begin{subfigure}[t!]{.5\columnwidth}
\includegraphics[draft=false,width=1.0\columnwidth]{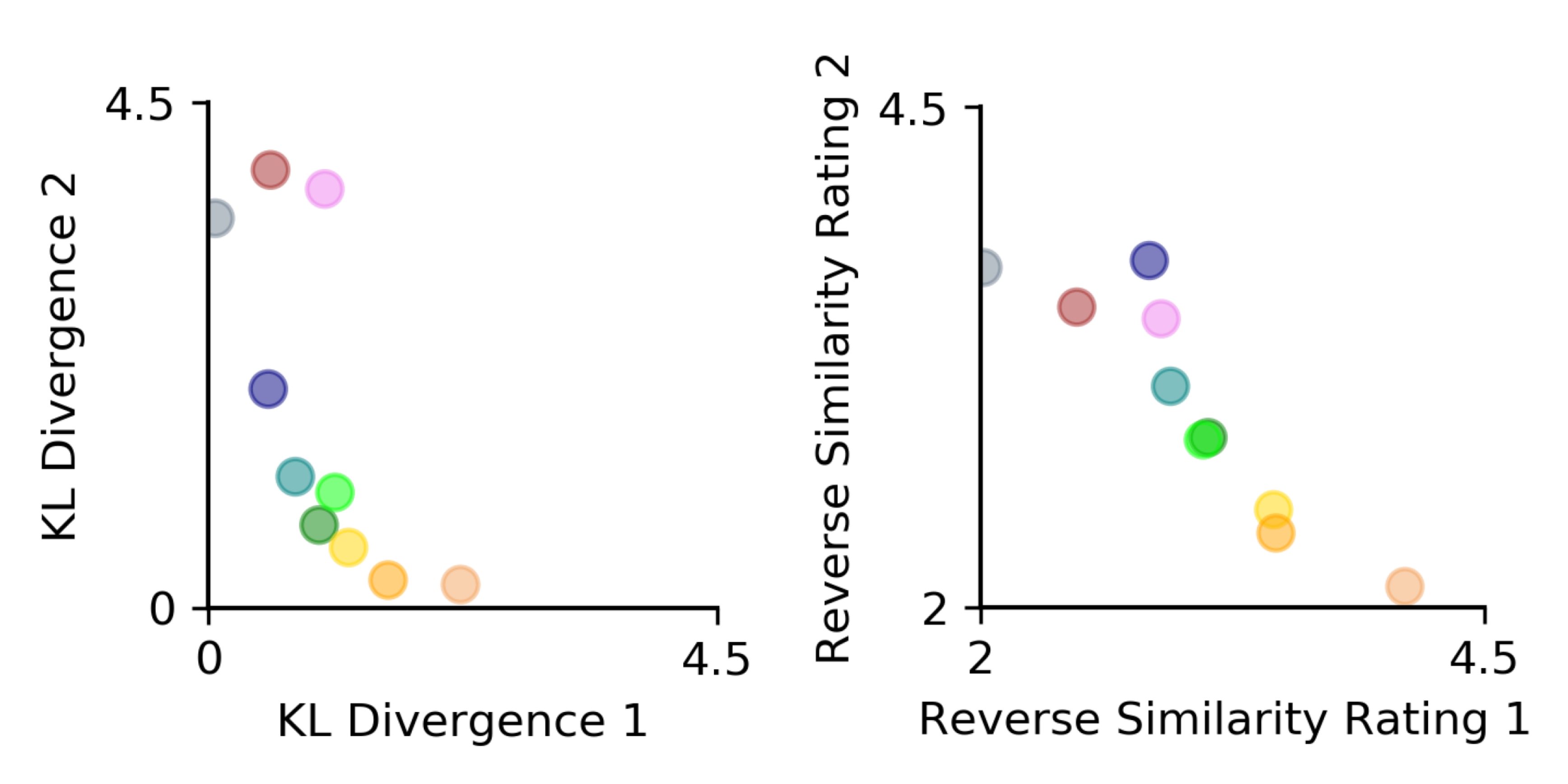}
%\caption{}
%\end{subfigure}\\
%\end{tabular}
\caption{ (Left) KL-divergence values of selected level scenes from the KL-divergence archive in Fig.~\ref{fig:map-elites-KL}-left. Each scene is assigned a unique color in the plot. (Right) Reverse mean ratings of similarity of the same scenes by participants. Level scenes that are identical between the two plots have the same color.}
\label{fig:study-similarity}
\vspace{-1em}
\end{figure}

\noindent\textbf{Similarity to Groundtruth Scenes.} We expect scenes with smaller KL-divergence values to be perceived as more similar to the two groundtruth scenes of Fig.~\ref{fig:KL-groundtruth}. In our second user study, we picked 10 scenes uniformly from the KL-divergence archive and presented them to participants.  

\textit{Dependent Measures.} We asked participants to rate how similar they considered the automatically generated scene to each of the two groundtruth scenes, on a Likert scale from 1 (very different) to 7 (very similar). 

\textit{Subject Allocation.} We recruited human participants through the Amazon’s Mechanical Turk service and followed the same selection process as in the previous study, resulting in 86 samples.

\textit{Analysis.} We fit two mixed-effects ordinal regression models to the data, one for each of the groundtruth levels, with the KL-divergence to that level as fixed effect and the participant id as random effect. The KL-divergence metric significantly predicted the perceived similarity for both groundtruth scenes ($\beta_1 = -1.24, t(773) = -10.97, p < 0.001$ and $\beta_2 = -0.40, t(773) = -8.84, p < 0.001$), supporting \textbf{H2}. Fig.~\ref{fig:study-similarity} contrasts the KL-divergence metrics of the selected scenes with their (reverse) mean ratings by the participants. While the two plots differ in scale, we see that the metrics capture well the perceived similarity.

% \begin{figure}[t!]
% \centering
% \includegraphics[draft=false,width=0.49\columnwidth]{figs/KL.png}
% \includegraphics[draft=false,width=0.49\columnwidth]{figs/responses.png}
% \caption{KL-Divergence values and reverse mean ratings of participants. The colors indicate different levels.}
% \label{fig:ratings}
%  \end{figure}

% \begin{figure}[t!]
% \centering
% \includegraphics[draft=false,width=0.66\columnwidth]{figs/difficulty.png}
% \caption{Mean ratings of perceived difficulty for the selected levels form the Representation-based archive. The radius of the markers is proportional to the rating.}
% \label{fig:ratings}
%  \end{figure}

% %Therefore, we conducted two user studies, where we asked to users to rate generated levels in terms of their difficulty (Representation-Based BCs) and 

\section{Conclusion}
We explored the use of QD algorithms to search the latent space of trained generator networks, to create content that has a diverse range of desired characteristics, while retaining the style of human-authored examples. In particular, we described an implementation where the QD algorithms MAP-Elites, MAP-Elites (line) and CMA-ME  were used to search the latent space of a DCGAN trained on level scenes from Super Mario Bros. In this problem, CMA-ME was superior to other tested algorithms in terms of coverage and QD-score, indicating that it finds a more diverse and high-quality set of level scenes.

QD algorithms extract a collection of scenes in a single run, rather than just one scene returned by optimization-based methods; their use is thus recommended when a collection of diverse, high-quality content is desired. We are excited about extending this work to search the latent spaces of other generative models, such as variational autoencoders~\cite{doersch2016tutorial} and generative pretraining models~\cite{chen2020generative}. Finally, we are excited about combining our approach with intelligent trial and error algorithms to create personalized levels~\cite{gonzalez2020finding}.
%QD algorithms extract a collection of levels from latent space in a single run; thus they can greatly improve sample efficiency 
%QD algorithms extract a collection of levels from latent space in a single run rather than just one level found in optimization-based methods, greatly improving the sample efficiency when a collection of levels is desired. Our method could also be used to evaluate and diagnose other kinds of PCGML methods that have control parameters akin to a latent vector.
%\input{8.body_conclusion.tex}
%\section*{Acknowledgements}
%We would like to congratulate the reviewers for finishing reading this paper.
\section*{Acknowledgements}
We would like to thank Sebastian Risi for his feedback on a preliminary version of this work.

\bibliography{bibliography.bib} 

%\begin{acks}
%Removed for anon review.
%\end{acks}

\end{document}